\documentclass[letterpaper, 10 pt, conference]{ieeeconf}  

\IEEEoverridecommandlockouts                              

\usepackage{cite}
\usepackage{comment}
\usepackage{amsmath, amssymb, amsfonts}
\usepackage{algorithm}
\usepackage{algpseudocode}
\usepackage{graphicx}
\usepackage{textcomp}
\usepackage{stfloats}
\usepackage{xcolor}

\def\BibTeX{{\rm B\kern-.05em{\sc i\kern-.025em b}\kern-.08em
    T\kern-.1667em\lower.7ex\hbox{E}\kern-.125emX}}
\begin{document}
\title{PACED-5G: Predictive Autonomous Control using Edge for Drones over 5G}
\author{Viswa Narayanan Sankaranarayanan$^{1}$, Gerasimos Damigos$^{2}$, Achilleas Santi Seisa$^{1}$, Sumeet Gajanan Satpute$^{1}$, \\ Tore Lindgren$^{2}$ and George Nikolakopoulos$^{1}$%
\thanks{This project has received funding from the European Union’s Horizon 2020 research and innovation programme under the Marie Skłodowska-Curie grant agreement No 953454.}
\thanks{$^{1}$ The authors are with the Robotics and AI Group, Department of Computer, Electrical and Space Engineering, Lule\aa\,\, University of Technology, Lule\aa\,\,}
\thanks{$^{2}$ The author is with Ericsson Research, Lule\aa\,\,}
\thanks{Corresponding Authors' email: {\tt\small (vissan, achsei)@ltu.se, gerasimos.damigos@ericsson.com}}
}
\maketitle
\begin{abstract}
With the advent of technologies such as Edge computing, the horizons of remote computational applications have broadened multidimensionally. Autonomous Unmanned Aerial Vehicle (UAV) mission is a vital application to utilize remote computation to catalyze its performance. However, offloading computational complexity to a remote system increases the latency in the system. Though technologies such as 5G networking minimize communication latency, the effects of latency on the control of UAVs are inevitable and may destabilize the system. Hence, it is essential to consider the delays in the system and compensate for them in the control design. Therefore, we propose a novel Edge-based predictive control architecture enabled by 5G networking, PACED-5G (Predictive Autonomous Control using Edge for Drones over 5G). In the proposed control architecture, we have designed a state estimator for estimating the current states based on the available knowledge of the time-varying delays, devised a Model Predictive controller (MPC) for the UAV to track the reference trajectory while avoiding obstacles, and provided an interface to offload the high-level tasks over Edge systems. The proposed architecture is validated in two experimental test cases using a quadrotor UAV.
\end{abstract}

\section{Introduction}
\label{sect:intro}
Unmanned Aerial Vehicles (UAVs) have an excellent capability to perform autonomous operations both indoors and outdoors. While autonomous UAVs are already in action for transportation tasks, industries have started to exploit the potential of UAVs for indoor and outdoor missions, such as inspection \cite{mansouri2018cooperative}, surveying, maintenance, and manipulation of power plants, pipelines, high-rise buildings, etc. (cf. \cite{6196930, 6820566, 7782757, 6761569, 7989609, 6878116, agha2021nebula, gupta2012survey}). In these operations, UAVs ensure human safety from operational hazards and ease out the operators' manual efforts. Furthermore, the global market is investing in expanding the capacities of UAVs to act as first responders in accidents and emergencies \cite{tomic2012toward} and to aid in the evacuation during disaster mitigation. Establishing autonomy in such critical applications is complex due to the unavailability of an accurate positioning system or a priori knowledge of the environment \cite{kanellakis2016evaluation}. Hence, most of these missions require complex exploration, mapping, state estimation, planning, control, and task allocation algorithms \cite{lindqvist2022compra} along with their intended tasks. However, executing all the tasks onboard increases the payload, reducing the UAV's flight time. An effective way to improve the performance and flight time of the UAV mission is to offload the heavy computational tasks to a remote system and run minimal control onboard.

For UAV control applications, several methods have been proposed with unique advantages (cf. \cite{sankaranarayanan2022adaptive, xu2018safe,  viswa2020asmc, ha2014passivity, sankaranarayanan2022survey}). In many cases, the UAV controller must incorporate a few constraints on the states \cite{Sankaranarayanan2022Robustifying, ganguly2021efficient} and control inputs \cite{lindqvist2020nonlinear} to include safety and optimality in power usage. Subsequently, Model Predictive Controller (MPC) \cite{lindqvist2020non} stands out as a solid contender to perform the UAV's desired tasks (such as trajectory tracking based on a high-level planner) while optimally ensuring multiple additional constraints. Offloading the MPC reduces the computational latency significantly, enhancing performance \cite{seisa2022edge}.

Offloading the control to a remote system introduces the additional challenge of latency in the loop. Edge computing that brings the storage and computational resources closer to the source of data enables the offloading to provide real-time processing of heavy algorithms with reduced latency \cite{shi2016edge}. The Edge-based control loops are further accelerated by the introduction of 5G networking, which offers high reliability with low communication delays \cite{hassan2019edge}. Hence, these systems are of research interest for remote operations that involve safety-critical processes. They have a wide range of applications, such as traffic management, healthcare services, autonomous vehicles, and smart homes \cite{varghese2016challenges, cao2020overview, tian2019adaptive, tian2019fog, tanwani2019fog}. Besides, deploying the individual modules inside independent containers optimizes the process and resources by various levels of security, abstraction, and resource-sharing methodologies \cite{pahl2015containers}. Notably, the Kubernetes (K8s) framework bundles sets of containers into groups (called PODs) to improve the real-time reliability, scalability, and security of the processes \cite{seisa2022edge, seisa2022comparison}. Many of the Edge-computing applications extensively use K8s in their architecture \cite{seisa2022edge, seisa2022cnmpc}.

Edge computing has received remarkable attention from the robotics community due to its potential use in applications such as Simultaneous Localization and Mapping (SLAM) of multiple agents \cite{sarker2019offloading}, teleoperation \cite{zhang2020toward}, object recognition \cite{barnawi2020intelligent}, exploration \cite{skarin2018towards}, and other industrial robotic applications \cite{chen2018industrial, spartharakis2020switching}. In particular, Edge-based remote offloading has enhanced the performance of MPCs with minimal latency, as presented in \cite{seisa2022anedge, seisa2022comparison, seisa2022cnmpc, seisa2022edge, aarzen2018control, skarin2020cloud, skarin2020cloud2, grafe2022event}. However, in a UAV application, even such a minimal closed-loop latency considerably deteriorates the performance \cite{sankaranarayanan2022adaptive}. Furthermore, none of these MPCs have considered delays in the control problem formulation. Given this premise, the contributions of this work are summarized in the following subsection.

\subsection{Contributions}
The main contribution of this work, Predictive Autonomous Control Using Edge for Drones over 5G (PACED-5G), is the development of an upgraded Edge-based control architecture for offloading high-level tasks in a remote environment that also handles ROS messaging and compensation for time-varying delays. Differently from \cite{seisa2022anedge, seisa2022cnmpc, seisa2022edge}, the upgraded architecture robustifies the system security over the network using a UDP tunneling approach. The dynamics of a UAV are modeled in the presence of delays in the system. A state estimator is designed to estimate the current state of the UAV using past observations and the available knowledge of the delay. The estimated delay in the network is updated using a moving average method to compensate for the time-varying nature of the delays. A nonlinear MPC is devised to ensure that the UAV follows the desired reference trajectory while avoiding obstacles and optimizing the control inputs. The proposed architecture exploits K8s orchestration for deploying the optimizer, MPC, state estimator, ROS master, trajectory generator, and UDP tunnel. The architecture is tested using a real-time quadrotor UAV, and the controller is validated using two experimental scenarios.

The rest of the article is organized as follows: The model of the system is derived along with the state predictor, and an MPC is formulated in Section \ref{sect:cont_dev}; Section \ref{sect:edge_arch} describes the Edge architecture; Section \ref{sect:exp_val} describes the experimental validation of the proposed control architecture, while Section \ref{sect:conc_futu} presents the conclusions and the potential future works.

\section{Controller Development}
\label{sect:cont_dev}
\subsection{UAV Modeling}
The overall delays in the closed loop can be accounted for in control by introducing a delayed input into the model of the UAV defined in \cite{kamel2017model}, given by,
\begin{align}
    \mathbf{\Ddot{p}}(t) &= \frac{1}{m} \mathbf{U}(t-\tau) + \mathbf{G} - \mathbf{A}\mathbf{\dot{p}}(t), \label{eq:pos_dyn} \\
    \dot{\phi}(t) &= \frac{1}{\alpha_\phi}(K_\phi \phi^d(t - \tau) - \phi(t)), \label{eq:phi_dyn} \\
    \dot{\theta}(t) &= \frac{1}{\alpha_\theta}(K_\theta \theta^d(t - \tau) - \theta(t)), \label{eq:theta_dyn} \\
    \mathbf{U}(t-\tau) &= \mathbf{R}(\mathbf{q}(t)) \begin{bmatrix}
        0 \\ 0 \\ F(t-\tau)
    \end{bmatrix}, \label{eq:control_mapping}
\end{align}
where $\mathbf{p} \triangleq \begin{bmatrix}
    x(t) & y(t) & z(t)
\end{bmatrix}^T \in \mathcal{R}^3$ and $\mathbf{q} \triangleq \begin{bmatrix}
    \phi(t), \theta(t), \psi(t)
\end{bmatrix}^T \in \mathcal{R}^3$ is the position is the orientation of the UAV in the inertial frame, $\mathbf{X_W - Y_W - Z_W}$, $m$ is the mass of the UAV, $\mathbf{U}(t)\in \mathcal{R}^3$ is the control input, $\mathbf{G}  \triangleq \begin{bmatrix}
    0 & 0 & -9.81
\end{bmatrix}^T$ is the gravity term, $\mathbf{A}\in \mathcal{R}^{3 \times 3}$ is a diagonal matrix with the drag-coefficients in its diagonals, ($\alpha_\phi, \alpha_\theta$) and ($K_{\phi}, K_{\theta}$) are time-constants and gains of inner-loop behaviors for roll and pitch respectively, ($\phi^d, \theta^d$) are the desired reference values for the roll and pitch angles, $\mathbf{R} \in \mathcal{R}^{3 \times 3}$ is the Euler angle rotation matrix, $F$ is the total thrust, and $\tau$ is the closed loop delay in the system. 

The UAV's linear and angular dynamics are partly decoupled with the position dynamics, dependent on the attitude of the UAV. So, the controller has a dual loop, where the inner loop controls the attitude dynamics, and the outer loop controls the position dynamics. Since the inner loop must run at a much higher frequency than the outer loop control to produce the desired moments, for the given control inputs, the inner loop control is placed on the onboard computer. Hence, the latency in the processing and actuation of the inner loop control is negligible. Since the outer loop control requires heavy computational power, it is offloaded on the Edge computer. The control inputs from the outer loop controller are the overall thrust, $F$, and the desired roll and pitch angles $\phi^d, \theta^d$. The yaw of the UAV is assumed to be zero at all times, since the reference input to yaw is always zero. Also, it is observed from the block diagram \ref{fig:arch} that the overall delay in the closed loop is given by $\tau = d_1 + d_2 + d_3$. 

Subsequently, the control problem is defined as the following:

\textbf{Control Problem:} Design a controller for a UAV to track a given trajectory in the presence of time-varying delays in the loop while avoiding the obstacles on the way.

\subsection{Proposed Control Solution}
The solution for the control problem is divided into two steps: State Estimator and Model Predictive Controller, which are further explained in the following subsections.

\subsubsection{State Estimator}
Since the inputs to the UAV are delayed by time, $\tau$, a state estimator has to be designed to predict the future estimate of the state for the current observation of the states based on the available information. The architecture is designed in such a way that the onboard PC loops back the time-stamped control input to the State Estimator along with the odometry of the UAV and the obstacles. The estimator finds an estimate of the closed-loop delay, $\widehat{\tau}$ in the network using the following formulation,
\begin{align}
    \widehat{\tau}(k+1) &= \widehat{\tau}(k) + \frac{(\tau_{new} - \widehat{\tau}(k))}{k+1}, ~ \tau(0) = 0, \label{eq:delay_est}
\end{align}
where $\tau_{new}$ is the difference between the current time-stamp and the time-stamp of the received control signal (delayed input).

The estimates of the position are defined as follows,
\begin{align}
    \mathbf{\widehat{p}}(t) &= \mathbf{p}(t-\tau), \label{eq:pos_est} \\
    \implies \mathbf{\dot{\widehat{p}}}(t) &= \mathbf{\dot{p}}(t-\tau). \label{eq:vel_est}
\end{align}
Since the control inputs are delayed, they have to be designed in such a way that the current inputs ensure that the future state after a delay tracks the future trajectory with minimal error. So, an estimate of future states of the UAVs and the obstacle need to be generated to be used in the control law. Using \eqref{eq:vel_est}, the future position of the UAV is estimated as, 
\begin{align}
    \mathbf{\dot{\widehat{p}}}(t + \tau) &= \mathbf{\dot{\widehat{p}}}(t) + \int_{t-\tau}^{t} \mathbf{\ddot{p}}(t) dt, \label{eq:vel_est_fut_1} 
\end{align}
The integral term in \eqref{eq:vel_est_fut_1} is simplified using a Taylor series approximation and ignoring the higher order terms (since $\tau ^2 << \tau$) as
\begin{align}
    \mathbf{\dot{\widehat{p}}}(t + \tau) &= \mathbf{\dot{\widehat{p}}}(t) + \mathbf{\ddot{p}}(t) \tau. \label{eq:vel_est_fut_2}
\end{align}
Using \eqref{eq:pos_dyn}, the expression \eqref{eq:vel_est_fut_2} can be expanded as
\begin{align}
    \mathbf{\dot{\widehat{p}}}(t + \tau) &= \mathbf{\widehat{p}}(t) + \left ( \frac{1}{m} \mathbf{U}(t-\tau) + \mathbf{G} - \mathbf{A}\mathbf{\dot{\widehat{p}}}(t+\tau) \right ) \tau, \nonumber \\
    \mathbf{\dot{\widehat{p}}}(t + \tau) &= (\mathbf{I} + \mathbf{A}\tau)^{-1} \left (\mathbf{\widehat{p}}(t) + \left ( \frac{1}{m} \mathbf{U}(t-\tau) + \mathbf{G} \right ) \right ) \tau
    \label{eq:vel_est_fut_3}
\end{align}
Further, future estimates of the position can be predicted from \eqref{eq:vel_est_fut_3} using the relationship,
\begin{align}
    \mathbf{{\widehat{p}}}(t + \tau) &= \mathbf{\widehat{p}}(t) + \mathbf{\dot{p}}(t) \tau. \label{eq:pos_est_fut}
\end{align}
The control input $U(t-\tau)$ depends on the attitude dynamics through the relationship, \eqref{eq:control_mapping}. So, the future estimates for the roll and pitch are derived by following similar steps from \eqref{eq:pos_est}-\eqref{eq:pos_est_fut}, and using the relationships \eqref{eq:phi_dyn}, \eqref{eq:theta_dyn}, as
\begin{align}
    \mathbf{\widehat{\phi}}(t) &= \mathbf{\phi}(t-\tau), \quad \mathbf{\widehat{\theta}}(t) = \mathbf{\theta}(t-\tau), \label{eq:att_est} \\
    \mathbf{{\widehat{\phi}}}(t+\tau) &=  \left (\mathbf{I} + \frac{\tau}{\alpha_{\phi}} \right )^{-1} \left (\mathbf{{\widehat{\phi}}}(t) - \frac{K_\phi}{\alpha_{\phi}}\phi_{d}(t-\tau) \right )\tau, \label{eq:phi_est_fut} \\
    \mathbf{{\widehat{\theta}}}(t+\tau) &=  \left (\mathbf{I} + \frac{\tau}{\alpha_{\theta}} \right )^{-1} \left (\mathbf{{\widehat{\theta}}}(t) - \frac{K_\theta}{\alpha_{\theta}}\theta_{d}(t-\tau) \right )\tau. \label{eq:theta_est_fut}
\end{align}

Similarly, the future estimate of the obstacle location, $\mathbf{p}*o$ is estimated as,
\begin{align}
    \mathbf{\dot{\widehat{p}}}^o(t+\tau) &= \mathbf{\dot{\widehat{p}}}^o (t) + \mathbf{G}\tau, \nonumber \\    
    \mathbf{{\widehat{p}}}^o(t + \tau) &= \mathbf{\widehat{p}}^o(t) + \mathbf{\dot{p}}^o(t) \tau, \label{eq:obs_est_fut}
\end{align}
where $\mathbf{\widehat{p}}_o(t) = \mathbf{p}_o(t-\tau)$ is the estimated position of the obstacle. Here, the considered obstacle has an initial velocity, but the only force acting on the obstacle is gravity. 

\subsubsection{Model Predictive Controller}
The objective of the high-level controller is to design the control inputs $\mathbf{u} \triangleq
\begin{bmatrix}
    F(t-\tau) & \phi^d(t-\tau) & \theta^d(t-\tau)
\end{bmatrix}^T$ to track the desired reference position trajectory that comes out of the trajectory generator node (cf. Fig. \ref{fig:arch}). Since the control inputs are delayed, they must be designed based on the estimated future states. So, a state vector is formed using the estimated future position, velocity, roll, and pitch, given by $\mathbf{x} \triangleq \begin{bmatrix}
    \mathbf{\widehat{p}}(t+\tau)^T & \mathbf{\dot{\widehat{p}}}(t+\tau)^T & \widehat{\phi}(t+\tau) & \widehat{\theta}(t+\tau)
\end{bmatrix}^T$ presented in Eq. \eqref{eq:vel_est_fut_3}-\eqref{eq:theta_est_fut}. The controller considers the state evolution of a specific number of time steps ($N$) with a sampling time, $T_s$, into the future (prediction horizon) to optimize the control inputs. The state evolution through the prediction horizon is obtained using the forward Euler method. The state evolution for the $(k+j)^{th}$ time step, predicted at the $k^{th}$ time step, is denoted by $\mathbf{x}_{k+j|k}$. The reference trajectory, $\mathbf{\Tilde{x}}^d$ is sampled through the prediction horizon for formulating the cost function, $J(\mathbf{\Tilde{x}}, \mathbf{\Tilde{u}}, \mathbf{u}_{k-1})$ as,
\begin{align}
    J &= \sum_{j=0}^N \{\left (\mathbf{x}^d_{k+j} - \mathbf{x}_{k+j|k} \right)^T \mathbf{Q}_x \left (\mathbf{x}^d_{k+j} - \mathbf{x}_{k+j|k} \right) \nonumber \\
    & + \left (\mathbf{u}_{k+j|k} - \mathbf{u}_{k+j-1|k} \right)^T \mathbf{Q}_{\delta u} \left (\mathbf{u}_{k+j|k} - \mathbf{u}_{k+j-1|k} \right) \nonumber\\
    & + \left ( \mathbf{u}_{k+j|k} + \mathbf{G} \right)^T \mathbf{Q}_u \left ( \mathbf{u}_{k+j|k } + \mathbf{G} \right)\}, \label{eq:mpc_cost}
\end{align}
where $\mathbf{x}^d$ is the reference state at each time step, $\mathbf{\Tilde{x}}, \mathbf{\Tilde{u}}$ are the appended states and inputs over the horizon, $\mathbf{Q}_x, \mathbf{Q}_u, \mathbf{Q}_{\delta u}$ are the positive definite gain matrices for states, inputs, and input rates, respectively. It is noticeable that the cost function not only minimizes the state errors but also ensures smoothness in the control signal by minimizing the difference between consecutive inputs through $\mathbf{Q}_{\delta u}$ and maintains the magnitude of the overall control inputs close to hovering mode through $\mathbf{Q}_u$.

Obstacle avoidance is imposed as an additional constraint over the cost function. Since the obstacles are considered to be unactuated objects, the Euler method state evolution is considered for their estimated positions over the prediction horizon. The obstacle is confined by a collision sphere of radius, $r_d$, and a spherical constraint of safety radius, $r_s$ is used to isolate the estimated states of the UAV from the estimated states of the obstacle over the prediction horizon, given by,
\begin{align}
     h^o &=  (r_s + r_d)^2 - \left (\mathbf{p}^o - \mathbf{p} \right)^T \left (\mathbf{p}^o - \mathbf{p} \right) \label{eq:obstacle}
\end{align}
The obstacle avoidance constraint in \eqref{eq:obstacle} is appended for all time steps of the prediction horizon. Additionally, bounding constraints are laid on the control inputs and the rate of desired roll and pitch to operate the system in a safe range on every $k^{th}$ time step over the complete horizon, given by,
\begin{align}
    \mathbf{u}^{max} &\leq \mathbf{u}_{k+j} \leq \mathbf{u}^{min}, \label{eq:control_constraint} \\
    -\delta \phi^{max} &\leq \phi^d_{k+j} - \phi^d_{k+j-1} \leq \delta \phi^{max},  \label{eq:phi_constraint} \\
    -\delta \theta^{max} &\leq \theta^d_{k+j} - \theta^d_{k+j-1} \leq \delta \theta^{max}, ~ \forall j \in \lbrace 1,2,...,N \rbrace \label{eq:theta_constraint}
\end{align}
where $\mathbf{u}^{max}, \mathbf{u}^{min} \in \mathcal{R}^{3 \times 3}$ are the minimum and maximum bounds of the control input, and $\delta \phi^{max}, \delta \theta^{max}$ are the maximum bounds on the rate of roll and pitch of the UAV.

Finally, the NMPC problem is solved in PANOC \cite{small2019aerial} using the following optimization objective,
\begin{align}
    \min_{\mathbf{\Tilde{x}}^K, \mathbf{\Tilde{u}}^K} J(\mathbf{\Tilde{x}}, \mathbf{\Tilde{u}}, \mathbf{u}_{k-1}) &\nonumber \\
    \quad \text{s.t.} ~ h^o_{k+j} > 0,& \nonumber\\
    \qquad \qquad \text{Constraints}  ~\text{\eqref{eq:control_constraint}} - \text{\eqref{eq:theta_constraint}}, & ~ \forall j \in \lbrace 0,1,...,N \rbrace. \label{eq:opt_obj}
\end{align}

\section{EDGE Architecture} \label{sect:edge_arch}

\begin{figure*}[!h]
    \centering
    \includegraphics[width=\textwidth]{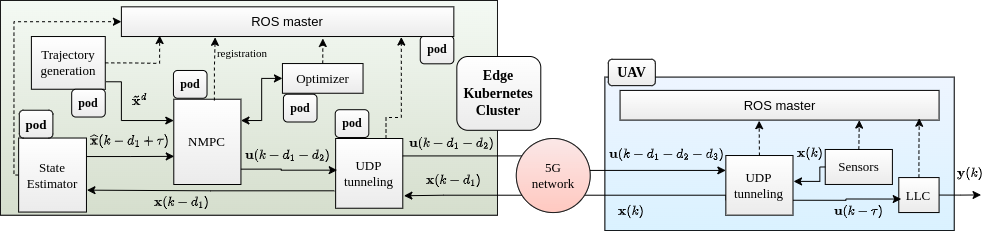}
    \caption{A schematic of the proposed Edge architecture}
    \label{fig:arch}
\end{figure*}

The Edge architecture handles the overall offboard computing required to control the UAV. In the proposed upgraded architecture, the control structure is divided into six ROS nodes: ROS master, MPC, Trajectory Generator, State Estimator, Optimizer, and UDP Tunnel. All the nodes run on their individual PODs, as shown in Figure \ref{fig:arch}. A Kubernetes cluster orchestrates all the PODs to form the remote system. Every POD has a single container that enables its functionality, whereas a container is a software unit that contains and runs specific application software and its libraries and dependencies. The containers are created using custom images based on ROS Noetic docker image entrypoints. 

Compared to the authors' previous work on Edge architectures (cf. \cite{seisa2022anedge, seisa2022edge, seisa2022cnmpc}), where a host network is used to enable the communication to and from the PODs, and thus, establish ROS messaging between the offboard and onboard computers, the upgraded architecture uses ROS messaging and UDP tunnelings over the 5G network for communicating between the offboard PODs and the onboard computers. Though the tunneling adds a small latency (of a few ms) in the loop, it eliminates the host network, which is unavailable in commercial Edge servers. Furthermore, using the host networks jeopardizes the system's security as it exposes the local services to the PODs, which an attacker could use to snoop on the network activity of other PODs or bypass the restrictive network policies in the namespace. The UDP tunnel has a server and a client. The UDP server subscribes to a ROS message and sends them over the 5G network as UDP packets, while the client receives the UDP packets from the network and publishes them as ROS messages.

The trajectory generator POD is used to run a high-level planner, which takes information from the world and generates a trajectory to be tracked by the UAV over a time horizon. The trajectory generator POD sends the reference trajectory, $\mathbf{\Tilde{x}}^d$, to the MPC POD using the reference topic. The MPC POD receives the estimated future states of the UAV, $\mathbf{\Tilde{x}}$ and the obstacle from the state estimator, $\mathbf{\Tilde{p}}^o$ using the UAV pose and obstacle pose topics. It formulates the cost function using the reference trajectory. The MPC and optimizer communicate to generate the desired control inputs ($F, \phi^d, \theta^d$) for the low-level onboard controller using the references, estimated UAV and obstacle states, and cost function. The MPC POD generates a time stamp before the optimization and uses the time stamp to publish the control inputs through the command velocity topic.

The UDP server on the Edge side subscribes to the command velocity topic and forwards them to the network as UDP packets (byte stream). The UDP client of the onboard side receives these packets and publishes them with and without time stamps. The command velocity topic without the time stamp is subscribed by the UAV's internal controller, while the one with the time stamp is looped back with the same time stamp on the old command velocity topic through a UDP server. The UDP server also subscribes to and transfers the odometry topics of the UAV and obstacles. The UDP client on the Edge side receives the UDP packets and publishes the information as odometry topics and the old command velocity topic.

The state estimator subscribes to the odometry and old command velocity topics. It uses the difference between the current time stamp and the time stamp in the old command velocity to predict the approximate delay, ${\tau}$ in the network. The odometry ($\mathbf{p}, \mathbf{\dot{p}}, \mathbf{p}^o, \mathbf{\dot{p}}^o$), estimated delay ($\widehat{\tau}$), and the old control inputs ($F, \phi^d, \theta^d$) are used to estimate the future state of the UAV, $\mathbf{\widehat{p}}, \mathbf{\dot{\widehat{p}}}$. Similarly, the delay ($\tau$) and odometry of the obstacles ($\mathbf{p}^o, \mathbf{\dot{p}}^o$) are used to estimate the future state of the obstacle, $\mathbf{\widehat{p}}^o$. The state estimator publishes the estimated future states of the UAV and the obstacles. A ROS master node runs on an independent POD, so any node failure would not affect it. All the nodes in the POD are registered to the ROS master node.

\section{Experimental Validation} \label{sect:exp_val}

\begin{figure}[!h]
    \centering
    \includegraphics[width=0.48\textwidth]{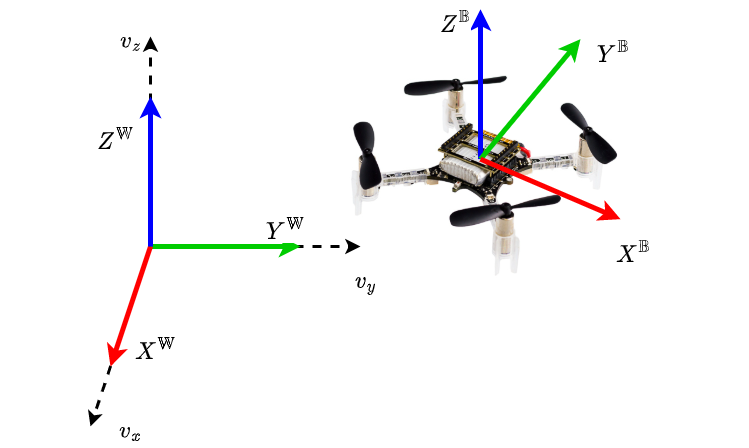}
    \caption{A schematic of the UAV used in the experiment.}
    \label{fig:cf}
\end{figure}

The performance of the PACED-5G architecture is validated experimentally using a quadrotor UAV, Crazyflie 2.0 Nano Quadcopter (cf. Fig. \ref{fig:cf}). The odometry feedback for the UAV and obstacle is obtained using a Vicon Motion Capture System. The Edge Architecture is deployed on a remote server at the datacenters of RISE RESEARCH INSTITUTES OF SWEDEN \cite{rise}, in Lule\aa\,\,. The specifications of the Kubernetes cluster are presented in Table. \ref{tb:cluster}. The edge datacenters provide significant computational power For the tracking demonstration, the trajectory generator provides a reference trajectory ($x^d(k) = sin(k/600) ~m, ~ y^d(k) = cos(k/600) ~m, ~ z(k) = 0.8 ~m$) to the MPC. The states are sampled at 30 Hz ($T_s \approx 0.033 ~s$) inside the MPC with a prediction horizon, $N = 60$ time steps. The experiment is performed in two scenarios, which are explained in the following subsections.

\begin{table}[htbp]
	\centering
	\caption{Kubernetes Cluster Specifications}

	\begin{tabular}{ c c } 
 	\hline
        \hline
 	\centering
 	\shortstack{\\Kubernetes Version} & v1.20.15\\
 	\hline
 	\centering
   	\shortstack{\\Worker Nodes} & 3 nodes\\
 	\hline
 	\centering
 	\shortstack{\\CPU} & 60 cores (20 per node)\\
 	\hline
 	\centering
 	\shortstack{\\Memory} & 590 GiB (196 GiB per node)\\
 	\hline
 	\centering
 	\shortstack{\\PODs for the application} & 6\\
 	\hline
	\end{tabular}
	\label{tb:cluster}
\end{table}

\subsection{Scenario 1: Comparison with uncompensated control}

\begin{figure}[!h]
    \centering
    \includegraphics[width=0.48\textwidth]{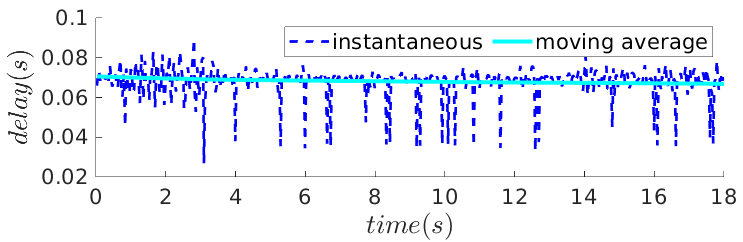}
    \caption{Instantaneous and moving average estimate of the delay in the closed loop over the flight duration.}
    \label{fig:delay}
\end{figure}

\begin{figure}[!h]
    \centering
    \includegraphics[width=0.48\textwidth]{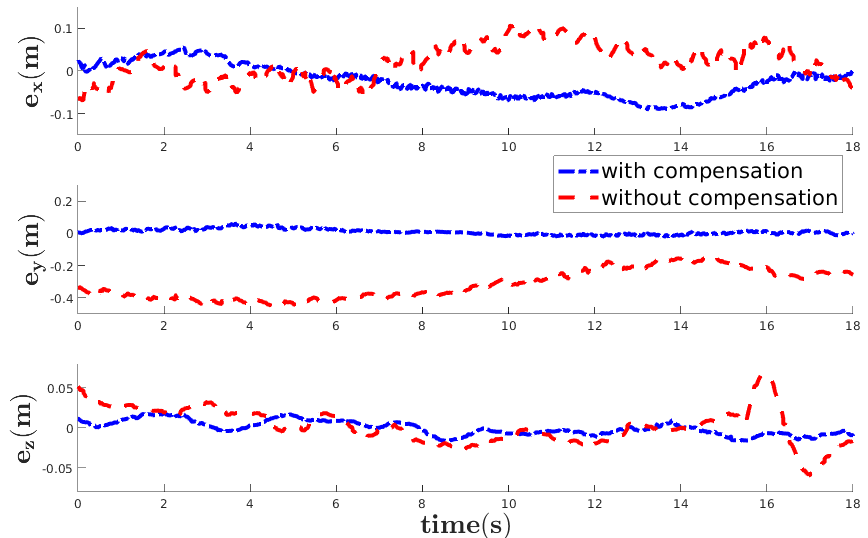}
    \caption{Tracking error in UAV's position}
    \label{fig:pos_err}
\end{figure}

\begin{figure}[!h]
    \centering
    \includegraphics[width=0.48\textwidth]{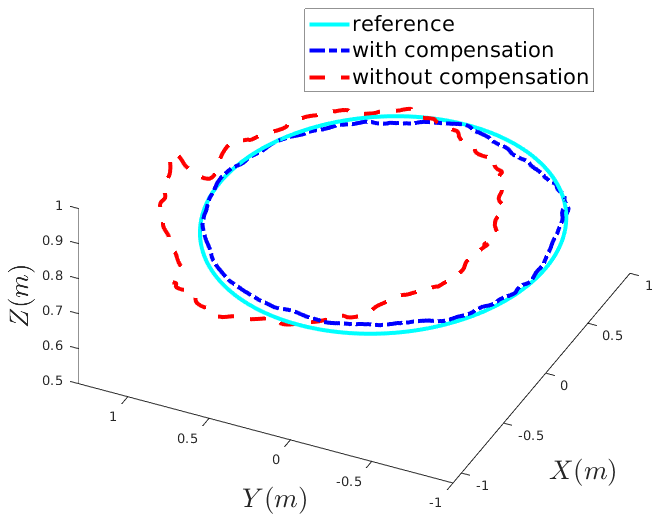}
    \caption{Trajectory tracked by the UAV with and without compensation for the delays in the system for the given reference trajectory.}
    \label{fig:traj}
\end{figure}

\begin{figure}[!h]
    \centering
    \includegraphics[width=0.48\textwidth]{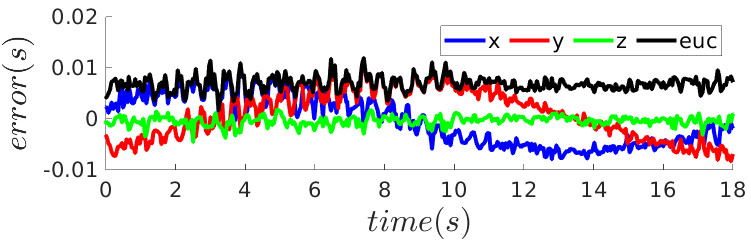}
    \caption{Error in the estimation of the states ($x, y, z$) and their Euclidean distance.}
    \label{fig:pred_err}
\end{figure}

\begin{table}[!t]
\renewcommand{\arraystretch}{1.1}
\caption{{Position Tracking Performance}}
\label{tb:rms_est_pos}
		\centering
{
{	\begin{tabular}{c c c c c}
		\hline
		\hline
		Controller & \multicolumn{4}{c}{Root mean squared error}  \\ \cline{1-5}
		 & $x(cm)$ & $y(cm)$  & $z(cm)$ & $Euc (cm)$  \\
		 \hline
		with estimator & 4.46  &  3.75  &  0.90 & 5.70 \\
		 \hline
		without estimator & 6.45  &  32.99  &  2.28 & 33.55   \\
		 \hline
          \hline
		estimation error & 0.49 & 0.50 & 0.11 & 0.69  \\
		\hline
\end{tabular}}}
\end{table} 

In this scenario, the need for the state estimator is presented by comparing the control of UAV with and without the state estimator. Fig. \ref{fig:delay} - \ref{fig:pred_err} highlight the experiment's results. The time-varying nature of the delays is observed in Fig. \ref{fig:delay}. The moving average delay estimate lies close to $67$ milliseconds over the entire duration. Though the architecture reduces the overall latency, the minimal delays in the system are sufficient to degrade the tracking performance. Fig. \ref{fig:pos_err} shows that the tracking error in position is significantly reduced using the state estimator. The same is observed in Fig. \ref{fig:traj}, which shows the 3D plot of the trajectories taken by the UAV with and without compensation. Without the estimator, the trajectory of the UAV is off to a large extent. However, the estimator noticeably improves the performance and maintains the actual trajectory quite close to the reference. Further, Fig. \ref{fig:pred_err} shows the effectiveness of the state estimator, where the error in the estimation (the difference between the actual value and the estimated value) is plotted. It is evident that the Euclidean distance between the estimated value and the actual value is less than $2$ millimeters at all the time steps.

The numerical analysis of the results is presented in Table. \ref{tb:rms_est_pos}. The Root Mean Squared (RMS) value of the errors in position shows that without an estimator, the deviation in performance, especially in the $y$ axis, is drastic, making it unsuitable for practical applications. However, the estimator minimizes the tracking error into a small margin, with the mean of the Euclidean distance between the reference and the actual trajectories less than 6 cm. Further, the RMS value of the estimation error is less than or equal to $0.5$ cm in all directions, with the mean Euclidean distance between the estimated value and the actual value being less than $7$ cm, hence, proving the reliability of the proposed state estimator in the control structure.

\subsection{Scenario 2: Obstacle Avoidance}

\begin{figure}[!h]
    \centering
    \includegraphics[width=0.48\textwidth]{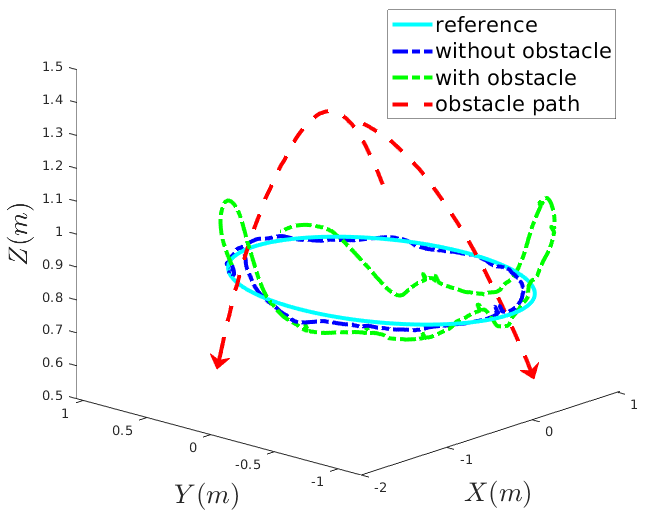}
    \caption{Trajectory taken by the UAV in the absence and presence of the obstacle while following a reference trajectory, along with the obstacle trajectories during possible collisions.}
    \label{fig:obs_traj}
\end{figure}

In this scenario, the UAV is commanded to track the trajectory in the presence of a moving obstacle. The obstacle is thrown at the drone in two instances when the UAV is following the commanded circular trajectory. Since the drone would not avoid the obstacle without the state estimator, which would damage the drone, this scenario is performed only for the case with the estimator. Fig. \ref{fig:obs_traj} shows the trajectory taken by the drone in the absence and presence of the obstacle. The NMPC predicts the possible future collision and moves the UAV away from the point of a collision immediately to maintain a safe distance from the obstacle, and brings the UAV back to the trajectory when the obstacle is avoided. Thus, the PACED-5G control architecture proves its dynamic obstacle avoidance capability in a delayed environment.

\section{Conclusions and Future Work} \label{sect:conc_futu}
An upgraded Edge architecture, PACED-5G, is proposed for UAVs to offload computationally heavy high-level control algorithms to a remote workstation. The utilization of ROS messages for the internode communication and the introduction of UDP tunneling to establish communication between the onboard and offboard computers over any existing 5G network removes the requirement on a host network. Further, it acts as a security layer protecting the Edge server from possible attacks. A state estimator is developed to compensate for the effects of time-varying closed-loop delays in the system. A nonlinear MPC is designed to follow any reference trajectory provided by the trajectory generator in the delayed environment. The architecture is tested on UAV hardware in two experimental scenarios. The results of the experiments are analyzed to show the efficacy of the proposed architecture.

The proposed controller can compensate for time-varying latency (moving average $67$ milliseconds). However, some interesting future direction would be to provide more redundancy to the system. If, for example, the communication is degraded or lost entirely, safety actions should be considered and onboard backup planning implemented. Furthermore, the utilization of edge resources can provide an environment through which multiple robots would be able to communicate and collaborate in order to execute more complex and demanding tasks, while Kubernetes would manage the resources and orchestration of the applications.

\bibliographystyle{IEEEtran}
\bibliography{references}

\end{document}